\documentclass[11pt]{article}

\usepackage[preprint]{acl}

\usepackage{times}
\usepackage{latexsym}
\usepackage{booktabs}
\usepackage[T1]{fontenc}

\usepackage[utf8]{inputenc}

\usepackage{multirow} 
\usepackage[table]{xcolor} 
\usepackage{microtype}

\usepackage{amsmath,amssymb} 
\usepackage{algorithm}      
\usepackage{algpseudocode}
\usepackage{inconsolata}

\usepackage{graphicx}

\title{Beyond Layer Importance in Layer-wise Sparsity: An Inter-Layer Perturbation-Absorption Perspective}

\author{Tao Jing, Ningxin Wu, Chen Kang, Dong Yu, Changliang Li, Pengyuan Liu}

\begin{document}
\maketitle
\begin{abstract}
The considerable layer-wise redundancy in large language models (LLMs) has established non-uniform sparsity allocation across layers as the standard pruning approach for efficient compression.
Existing layer-wise allocation methods that estimate allocation strategy from local signals such as activation outliers or weight spectra mainly derive from local layer importance, whereas the final post-pruning performance is also influenced by the network's subsequent compensatory capacity. In this paper, we directly characterize this property through controlled perturbation experiments. We make the following empirical findings. First, layers exhibit highly heterogeneous responses to pruning-scale perturbations. In most cases, early layers amplify perturbations, while middle and late layers actively absorb them, with relative L2 drift decreasing monotonically across depth and direction realigning toward the unperturbed hidden-state trajectory. Second, absorption is a large-perturbation phenomenon. Under small perturbations the network exhibits amplification across all layers, and the transition to absorption occurs smoothly as perturbation magnitude grows to pruning scale. This enriches the linearized accumulation theory underlying related works. Building on these findings, we define an absorption coefficient per layer and propose absorption-aware correction, an orthogonal augmentation that improves OWL and AlphaPruning by reducing perplexity by 7.13\% and boosting zero-shot accuracy by 1.02\% across multiple model families at 70\% sparsity.

\end{abstract}

\section{Introduction}

Large Language Models (LLMs) have achieved remarkable capabilities in various natural language processing tasks \citep{liu2024deepseek, touvron2023llama, achiam2023gpt}. Scaling up model parameters consistently brings performance improvements \citep{wei2022emergent}, yet their massive computational and memory footprint present substantial obstacles to model deployment, inhibiting wider applicability\citep{wang2024model}. 

Pruning has emerged as a key approach for compressing large language models (LLMs) without retraining, with methods such as SparseGPT \citep{frantar2023sparsegpt} and Wanda \citep{sun2024simple}. A central design choice in post-training pruning is layer-wise sparsity allocation: how to distribute a global sparsity budget across the layers of the network. Uniform allocation discards the fact that layers differ in their sensitivity to pruning, and a growing line of work has therefore proposed non-uniform allocation strategies.

Existing approaches can be organized into three families. Metric-based methods estimate a per-layer importance score from a hand-crafted signal and allocate sparsity inversely to importance. OWL \citep{yin2024outlier} uses activation outlier ratios, AlphaPruning \citep{lu2024alphapruning} uses heavy-tailed self-regularization theory applied to weight spectra, and ALS \citep{li2024adaptive} uses mutual information between layer activations. Search-based methods instead treat allocation as an optimization problem, where DSA \citep{li2024discovering} and EvoPress \citep{sieberlingevopress} both use evolutionary search over allocation expressions.
A third family explicitly considers inter-layer coupling. ATP \citep{huang2025determining} derives a layer-wise allocation from a theoretical analysis of how reconstruction error propagates forward through the network, proving that increasing the sparsity of an earlier layer raises the lower bound on the reconstruction error of all subsequent layers, and prescribing a monotonically increasing arithmetic progression as the remedy.

Metric-based methods mainly rely on layer-intrinsic signals and therefore overlook inter-layer coupling. ATP incorporates such coupling by modeling how pruning a layer affects downstream behavior through the linear propagation of reconstruction error. However, this formulation neglects the network's actual response to pruning-induced perturbations. Since pruning a layer introduces perturbations that propagate through downstream layers and may be either amplified or absorbed, layer-wise pruning should account for how the downstream network processes the perturbation caused by each pruned layer.

In this paper, we directly study this downstream response by using controlled layer-wise perturbation injection to trace how perturbations evolve across depth in a transformer, yielding insights that motivate a new view of layer-wise allocation.

\textbf{Layer-dependent response.} Under pruning-scale perturbations, layers exhibit highly heterogeneous responses. Typically, perturbations injected into early layers are commonly amplified by subsequent layers; perturbations injected into middle and late layers are actively absorbed, and their direction realigns toward the unperturbed hidden-state trajectory. The strength of absorption varies substantially across layers and is not well predicted by depth alone.

\textbf{Scale-dependent response.} Absorption is a large-perturbation phenomenon. Under small perturbations, all layers exhibit accumulation: relative drift grows monotonically with depth regardless of injection point. As the perturbation magnitude increases toward the scale produced by actual pruning, middle and late layers smoothly transition into the absorbing regime. The transition is continuous, ruling out numerical artifacts. 


Taken together, these findings do not contradict the observations of reconstruction error accumulation in prior work. The growth of total error along depth is the cumulative superposition of many local perturbations, and the error induced by each local perturbation may be absorbed or amplified as it propagates through the rest of the network. Our results identify this downstream response as a missing dimension of layer importance.

To verify that this dimension is practically useful, we propose a simple operationalization. We define a per-layer drift ratio and use it as an additive correction on top of metric-based allocation methods, redistributing a small fraction of the sparsity budget while keeping the global budget fixed. Our intent is not to claim a new state-of-the-art method but to demonstrate that the absorption dimension translates into measurable gains. The results of experiments show that the correction improves OWL and AlphaPruning in most settings.

The central contributions of this paper can be summarized as follows.

\begin{itemize}
    \item We establish that pruning-induced perturbations are subject to partial absorption by the rest of the network. We further show that absorption is layer-dependent and scale-dependent.
    \item We provide a precise relationship between our findings and the linearized accumulation theory. The depth-wise growth of total error under whole-model pruning is the cumulative effect of many partially-absorbed local perturbations, not the amplification of every single one.
    \item We define a per-layer drift ratio and use it as a correction on top of metric-based allocation methods, showing improvements across several models as our analysis predicts.
\end{itemize}

\section{Related Work}

\noindent \textbf{LLM Pruning.}
Pruning has long been used to compress neural networks, dating back to magnitude-based and second-order criteria such as OBD \citep{lecun1989optimal} and OBS \citep{hassibi1992second}. For LLMs, the prohibitive cost of retraining has motivated post-training unstructured pruning, where weights are removed in a single pass with only a small calibration set. SparseGPT \citep{frantar2023sparsegpt} selects masks and updates weights using inverse-Hessian information; Wanda \citep{sun2024simple} prunes weights with the lowest product of magnitude and input activation norm; further refinements include ALPS \citep{meng2024alps}. These methods determine which weights to remove within each layer but typically apply a uniform sparsity rate across layers, leaving the question of how to allocate sparsity across layers untouched. Our work focuses on this layer-wise allocation problem and is compatible with any of the underlying pruning methods above.

\noindent \textbf{Non-Uniform Sparsity.}
Recent work allocates sparsity non-uniformly across layers, motivated by the observation that layers contribute unequally to model performance \citep{frankle2018lottery}. These methods fall into three families. Metric-based methods assign each layer an importance score from a hand-crafted signal. OWL \citep{yin2024outlier} and MRP \citep{gao2025maximum} uses activation outliers, AlphaPruning \citep{lu2024alphapruning} uses heavy-tailed weight spectra, and LSA \citep{yanglsa} uses each layer's minimal linear reconstruction error. Search-based methods such as DSA \citep{li2024discovering} instead optimize allocation directly at the cost of repeated trial prunings. Methods considering inter-layer coupling: ATP \citep{huang2025determining} is a prior method that explicitly models how pruning at one layer affects others, deriving a monotonically increasing arithmetic progression from a linearized analysis of reconstruction error propagation. None of the above directly characterizes how a pruning-induced perturbation propagates through downstream layers in the regime where pruning actually operates, a dimension we identify in Section~\ref{sec:method} as the key missing factor.

\noindent \textbf{Reconstruction Error in Pruning.} Reconstruction error has long served as both an objective and a signal in LLM compression, and lower reconstruction error typically indicates that the compressed network retains the performance of the original network more faithfully \citep{yun2021all, luo2017thinet, he2017channel, zhuang2018discrimination, meng2024osscar}. SparseGPT \citep{frantar2023sparsegpt} selects masks and updates weights to minimize per-layer reconstruction error; LSA \citep{yanglsa} uses the minimal linear reconstruction error of each layer as an importance signal; ATP \citep{huang2025determining} builds its allocation theory on the propagation of reconstruction error across depth. In this paper, we explore the reconstruction error compensation phenomenon in LLMs. Specifically, the reconstruction error is continuously repaired during propagation through subsequent layers, and different layers exhibit varying perturbation recovery capabilities. The non-uniform pruning of the model should take this difference into account and accordingly adjust the sparsity rate allocation for each layer. Therefore, we propose a reconstruction error compensation factor to quantify the perturbation recovery capability of different layers.

\begin{figure*}[t]
    \centering
    \includegraphics[width=\textwidth]{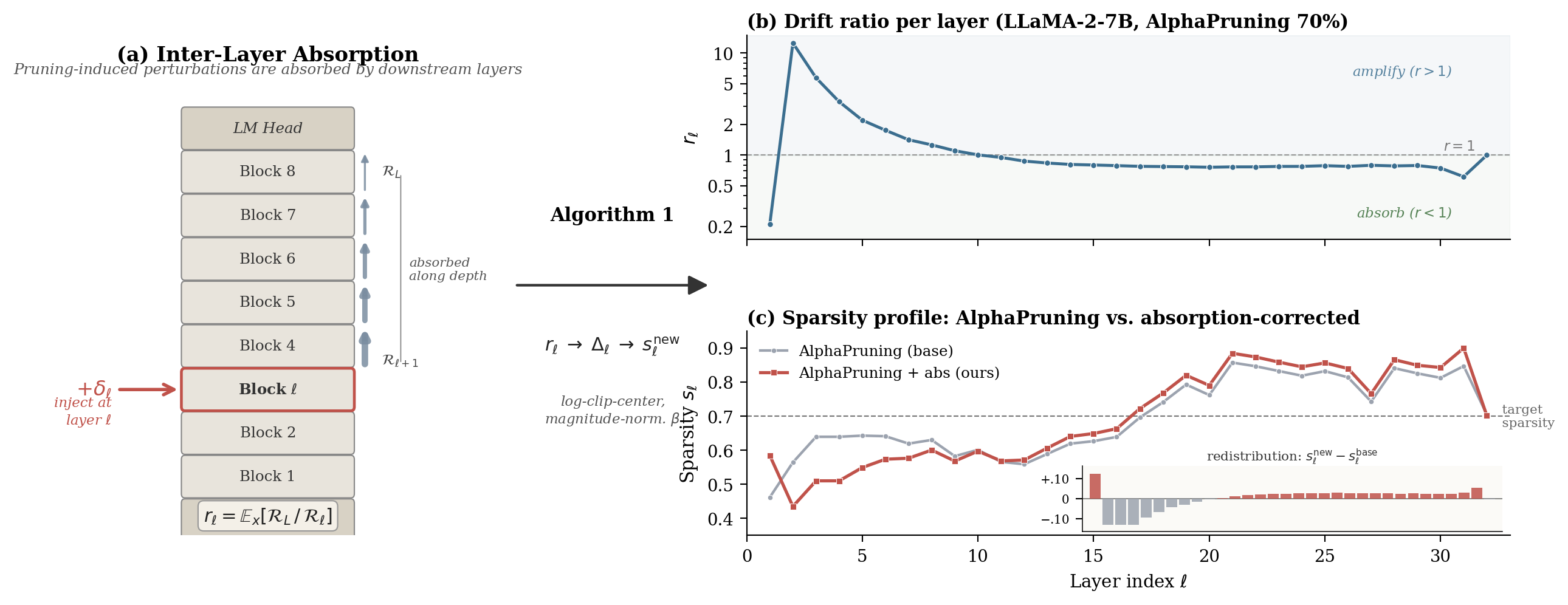}
    \caption{\textbf{Inter-layer perturbation absorption and the absorption-aware correction it enables.} \textbf{(a)} a perturbation $\delta_{\ell}$ injected at layer $\ell$ propagates through downstream layers, and in most cases its magnitude $\|\Delta h_k\|$ decays along depth at pruning scale. This is the inter-layer absorption phenomenon we identify in Section \ref{sec:3.3}. The end-to-end drift ratio $r_\ell = \mathbb{E}_x \left[ \frac{\mathcal{R}_L}{\mathcal{R}_\ell} \right]$ summarizes this per-layer behavior. \textbf{(b)} Measured on LLaMA-2-7B after AlphaPruning at 70\% sparsity (Gaussian probe $\eta = 10^{-2}$, 5 seeds), the figure shows the layer-wise drift ratio. \textbf{(c)} Algorithm~\ref{alg:correction} converts $r_{\ell}$ into an additive correction on top of the AlphaPruning base allocation, strictly preserving the global sparsity budget. Sparsity is shifted away from layers with $r_{\ell}>1$ (amplification, dangerous to prune) toward layers with $r_{\ell}<1$ (absorption, safer to prune).}
    \label{fig:main}
\end{figure*}

\section{Layer-Dependent Reconstruction Error Compensation }
In this section, we design a controlled perturbation injection programme that isolates how a single layer's perturbation propagates through the rest of the network, then use it to arrive our findings about how downstream transformer layers respond to perturbations.
\label{sec:method}
\subsection{Preliminaries}
We treat each layer of a large language model (LLM) as the elementary unit of study. Given an LLM with $L$ layers, the reconstruction error of the $\ell$-th layer ($l = 1, 2, \dots, L$) is formulated as

\begin{equation}
\mathcal{R}_\ell^{\text{abs}} \, = \big\| W_\ell X_\ell^T - (M \odot W_\ell)X_\ell^T \big\|_F^2
\label{eq:recon_abs}
\end{equation}

where $W_\ell \in \mathbb{R}^{c_{\mathrm{out}}\times c_{\mathrm{in}}}$ and 
$X_\ell \in \mathbb{R}^{N\times c_{\mathrm{in}}}$ denote the weight matrix and calibration inputs of layer $\ell$, respectively. 
Here $N$ is the number of calibration tokens, $M \in \{0,1\}^{c_{\mathrm{out}}\times c_{\mathrm{in}}}$ is the pruning mask, $M\odot W_\ell$ denotes the pruned weight matrix, and $\|\cdot\|_F$ stands for the Frobenius norm.

Existing post-training sparsity methods (e.g., SparseGPT) and non-uniform pruning strategies (e.g., LSA) commonly aim to minimize the reconstruction error of the pruned LLM.  
Despite its prevalent use, this metric carries an inherent limitation: it is scale-sensitive.
The magnitude of $\mathcal{R}_\ell^{\text{abs}}$ is directly influenced by the norm of hidden states, which grow substantially with depth in trained transformers.
Prior diagnostic studies have offered indirect evidence of this issue, observing that layers in depthwise-separable convolutional networks exhibit dramatically fluctuating dynamic ranges~\cite{yun2021all}.

To mitigate this, this work adopts a scale-invariant alternative. 
Let $h_\ell \in \mathbb{R}^{d \times n}$ denote the original output of layer $\ell$ on a batch of $n$ tokens, and $\widetilde{h}_\ell$ the corresponding output under intervention.
We define the relative reconstruction error of the layer $\ell$ as:

\begin{equation}
\mathcal{R}_\ell \,=\, \frac{\|\widetilde{h}_\ell - h_\ell\|_F}{\|h_\ell\|_F}.
\label{eq:recon}
\end{equation}

We use it together with the cosine distance $1 - \cos(\widetilde{h}_\ell, h_\ell)$ as a directional companion. With this metric modification, we further study the overall evolutionary characteristics of error propagation in pruned large language models and reveal the phenomenon of \textbf{reconstruction error absorption} in the next section.


\subsection{Perturbation Injection.} 
The reconstruction error has two sources, including the local error introduced by pruning a given layer, and the propagated error inherited from earlier pruned layers. To isolate propagation, we adopt layer-wise perturbation injection: at a chosen injection layer $\ell$, we replace its output with $\widetilde{h}_\ell = h_\ell + \delta_\ell$, leave all upstream layers untouched, and let the perturbed activation flow through the remaining layers $k > \ell$. We then record $\mathcal{R}_k$ at every subsequent layer. Two natural choices of injected $\delta_\ell$ exist. Isotropic Gaussian noise scaled to a target relative magnitude, and the actual activation difference produced by pruning the layer's weights. The two probes serve different purposes, and we use Gaussian noise scaled to match a target relative magnitude $\eta$: $\delta_\ell = \eta \cdot \|h_\ell\|_F \cdot \xi$, where $\xi$ is a unit-norm Gaussian tensor. 

\subsection{Layer-Dependent Response.} 
\begin{figure}[ht] 
  \centering 
  \includegraphics[width=0.8\linewidth]{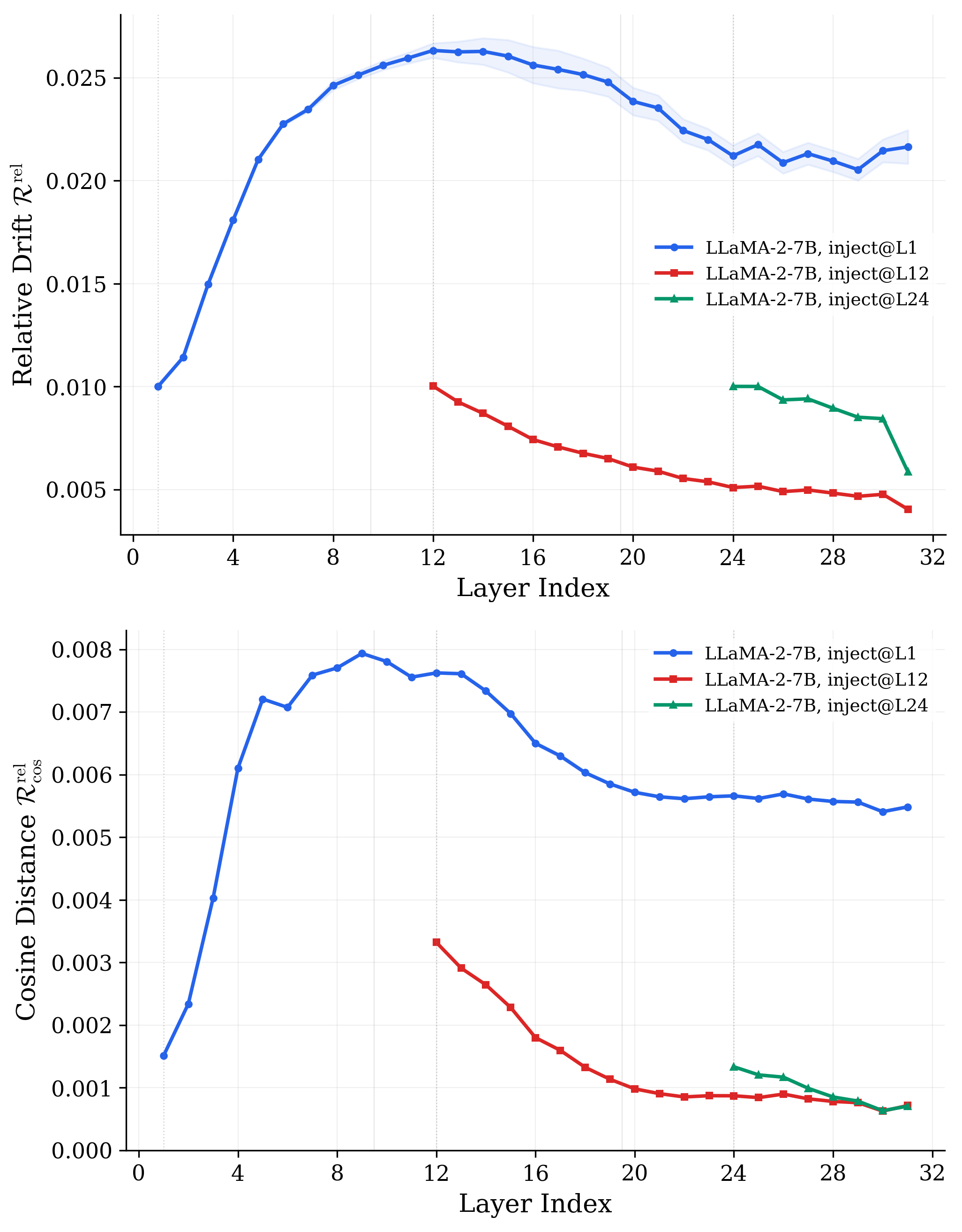} 
  \caption{Layer-dependent response of LLaMA-2-7B to perturbations. (Top) L2 relative drift $\mathcal{R}_k$ along depth, for Gaussian probes ($\eta = 10^{-2}$) injected at representative early, middle, and late layers. (Bottom) Cosine distance between the perturbed and unperturbed hidden states for the same injections.} 
  \label{fig:l2_example} 
\end{figure}

We inject Gaussian perturbations of relative magnitude $\eta = 0.01$ at different layers and trace $\mathcal{R}_k$ across all subsequent layers $k > \ell$. The magnitude is chosen to match the scale produced by actual pruning. Figure \ref{fig:l2_example} presents the result for LLaMA-2-7B with injection at representative layers (full coverage of all 32 layers and additional models is provided in Appendix \ref{app:drift_all}).

As is shown, perturbations injected at some layers are amplified, while those injected at other layers are absorbed. Absorption capacity is heterogeneous across layers at similar depths. Layers separated by only a few positions in the middle of the network can exhibit substantially different absorption ratios. This rules out the possibility that absorption is captured by depth alone. Furthermore, the cosine trajectory shows that absorbing layers do not merely shrink the norm of the perturbation, but also actively rotate it back toward the unperturbed hidden state. The network is not passively damping a disturbance but re-mapping perturbed states onto trajectories close to its original behavior.

Since the perturbations caused by Gaussian noise differ from those produced by real pruning, there is a risk that the observed layer-dependent response is an artifact of probing with random directions. We address this concern in Appendix \ref{app:TBD}, demonstrating that the Gaussian probe is adequate for studying perturbation absorption.


\label{sec:3.3}

\subsection{Absorption Is a Large-Perturbation Phenomenon.} 
We next ask whether the layer-dependent response is a generic property of trained transformers, or whether it emerges only at certain perturbation magnitudes. Holding the injection layer and the random direction fixed, we sweep the relative magnitude $\eta$ across multiple orders of magnitude and re-measure $\mathcal{R}_k$ along depth.

\begin{figure}[ht] 
  \centering 
  \includegraphics[width=0.8\linewidth]{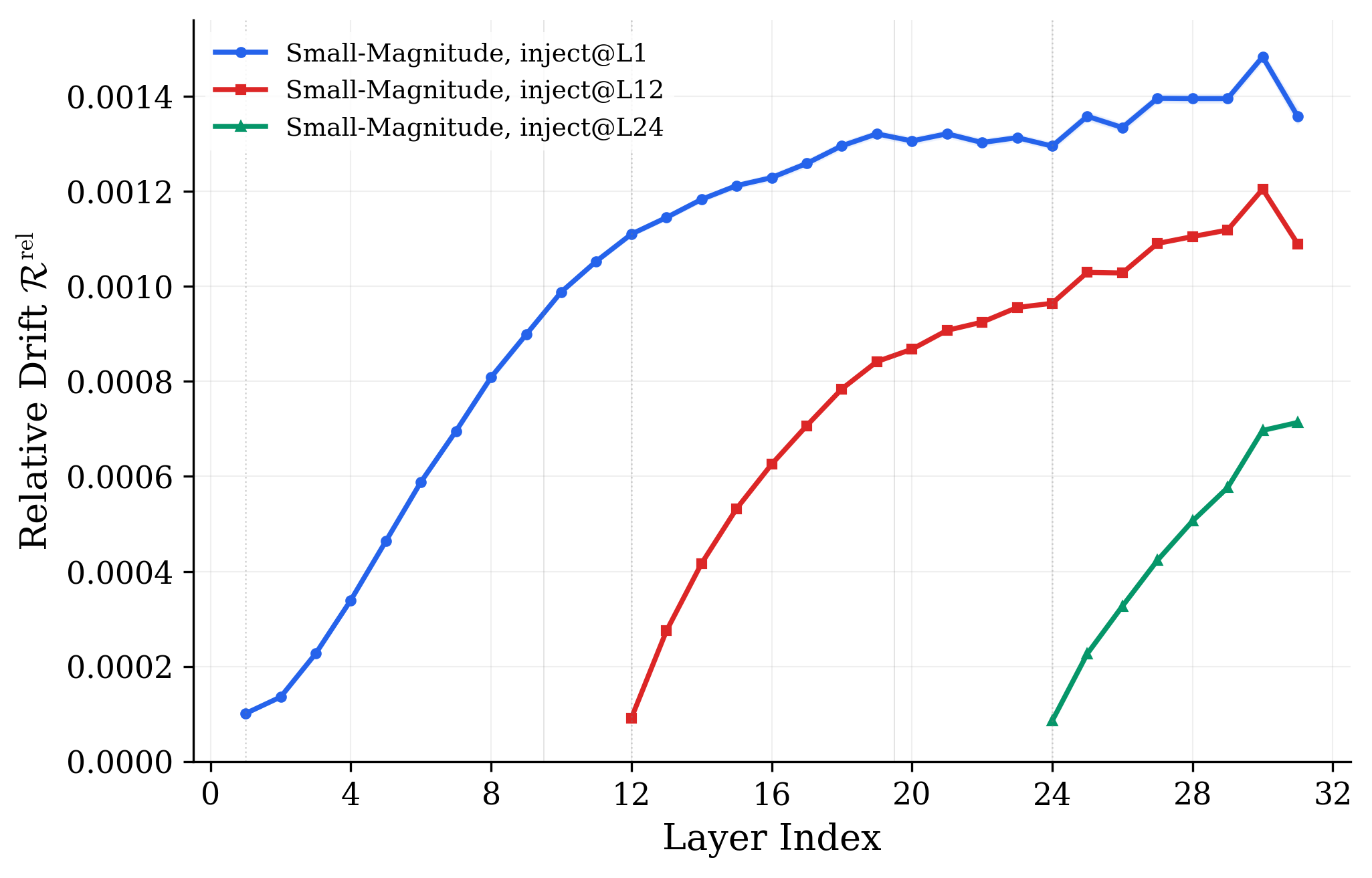} 
  \caption{Layer-dependent response of LLaMA-2-7B to perturbations. L2 relative drift $\mathcal{R}_k$ along depth, for Gaussian probes ($\eta = 10^{-4}$) injected at representative early, middle, and late layers. } 
  \label{fig:l2_small} 
\end{figure}

Figure~\ref{fig:l2_small} presents the result for LLaMA-2-7B with injection at layers 1, 12, and 24 for $\eta = 10^{-4}$. 
The two regimes, $\eta = 10^{-2}$ and $\eta = 10^{-4}$, are sharply distinct. At $\eta = 10^{-2}$, early-layer injection amplifies, middle and late-layer injection is absorbed, with $\mathcal{R}_k$ decreasing monotonically along depth. At $\eta = 10^{-4}$, every injection layer, including those that absorbed at pruning scale, instead exhibits that $\mathcal{R}_k$ grows monotonically along depth. The same network, the same direction, the same injection layer, and the qualitative behavior reverses with magnitude.

What remains as the most plausible mechanism is the nonlinear structure of the network. At small perturbation magnitudes, the network operates in its linearized regime, where each layer's effect on the perturbation is governed by its local Jacobian, and products of Jacobians with spectral norm exceeding one accumulate. At pruning scale, perturbations are large enough to engage the network's nonlinear components, namely LayerNorm normalization across features, attention softmax redistribution, and activation function saturation, which actively re-map perturbed states toward the manifold of hidden states the network was trained to produce.

\begin{figure}[ht] 
  \centering 
  \includegraphics[width=0.8\linewidth]{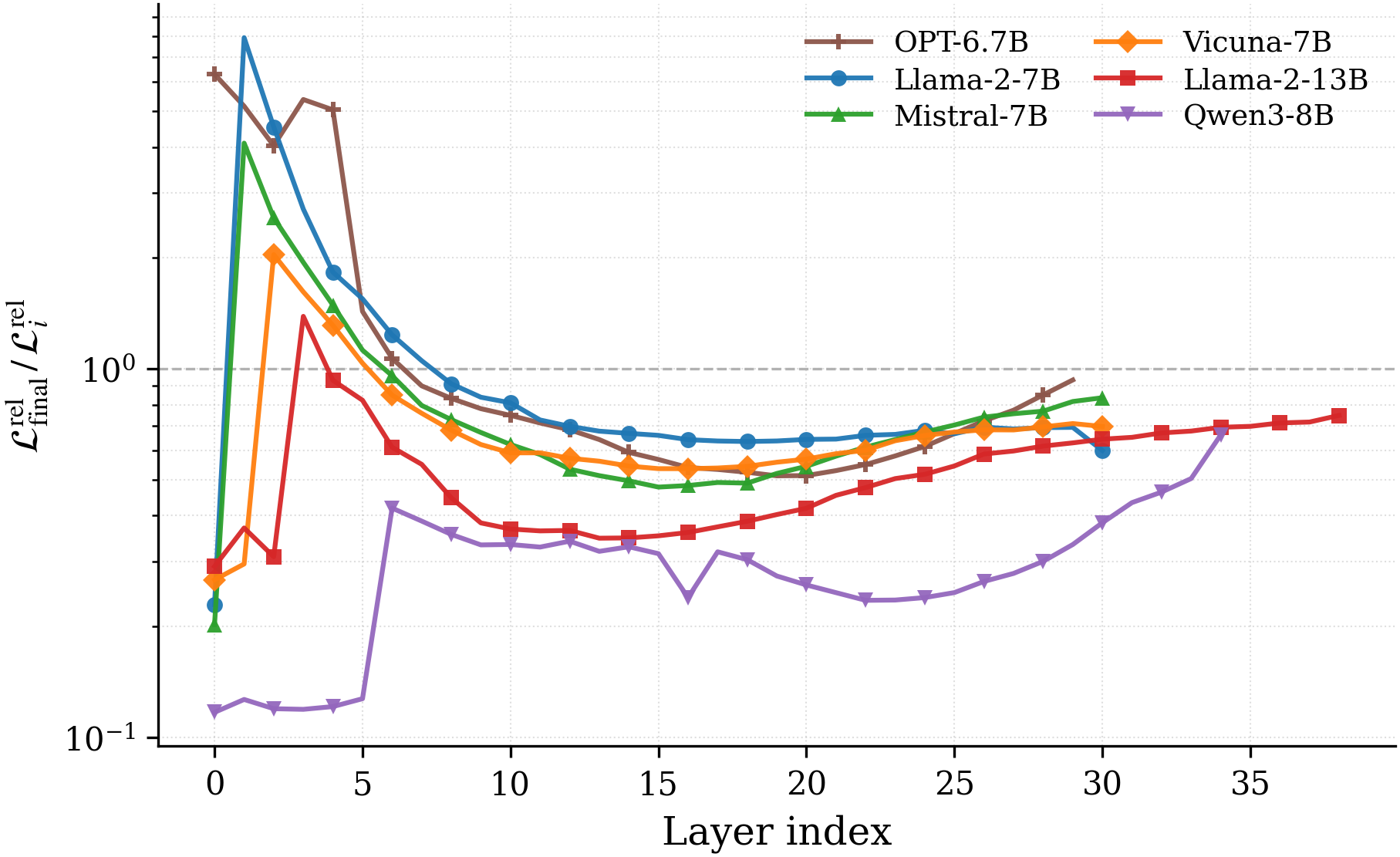} 
  \caption{Layer-dependent response across model families.} 
  \label{fig:multi_drift_ratio} 
\end{figure}

\subsection{Absorption Coefficient.}
For a layer $\ell$, we define the absorption coefficient as the ratio of final-layer relative drift to the relative drift at the injection point, averaged over a small calibration set:

$$
r_\ell = \mathbb{E}_x \left[ \frac{\mathcal{R}_L}{\mathcal{R}_\ell} \right],
$$

where
$L$ denotes the final transformer layer. Values $r_\ell < 1$ indicate absorption; values $r_\ell > 1$ indicate amplification.

Figure \ref{fig:multi_drift_ratio} visualizes the layer-wise absorption coefficient for multiple models. All these models exhibit the same overall layer-wise structure, and the magnitude and layer-wise smoothness of absorption differ noticeably across models. It indicates that $r_\ell$ captures architecture- and training-specific information that depth alone cannot. This cross-model consistency in structure, combined with model-specific variation in magnitude, is exactly the profile required of a useful allocation signal.

\label{sec:coefficient}

\section{Inter-Layer Coupling-Aware Allocation Correction.}

Layer-wise allocation faces an inter-layer coupling problem, where the cost of pruning a layer depends not only on the layer's own properties, but also on how subsequent parts of the network process the perturbation introduced by pruning, a problem that metric-based methods cannot address. Per-layer importance computed in isolation lacks consideration of inter-layer coupling. We therefore propose a correction method applied on top of metric-based methods, based on the drift ratio $r_{\ell}$ defined in Section~\ref{sec:coefficient}, to adjust the sparsity allocation of the metric-based approach.

\subsection{From Drift Ratio to Correction Signal.}
We now use the drift ratio $\{r_\ell\}$ to improve existing allocation methods. Given a profile $\{s_\ell^{\text{base}}\}$ produced by an existing method (OWL, AlphaPruning), we redistribute a portion of the sparsity budget across layers based on $\{r_\ell\}$, while strictly preserving the global budget $\sum_{\ell} s_{\ell} = \sum_{\ell} s_{\ell}^{\text{base}}$.

The raw drift ratio $r_\ell$ has three properties that make it unsuitable for direct use as an additive correction. It is multiplicatively scaled, it is non-negative but unbounded above, and it is contaminated by boundary effects. We address all three with a log-clip-center transformation:
\begin{equation}
\begin{split}
u_\ell &= \operatorname{clip}\bigl(\log r_\ell,\, -L_0,\, +L_0\bigr), \\
\Delta_\ell &= -\bigl(u_\ell - \bar{u}\bigr),
\end{split}
\label{eq:correction-signal}
\end{equation}
where $L_0 = \log 3$ and $\bar{u} = \tfrac{1}{L}\sum_\ell u_\ell$ is the mean of the clipped log-ratios.

\subsection{Applying the Correction.}
A naive choice for the correction strength would fix a scalar $\lambda$ and apply $s_\ell^{\text{new}} = s_\ell^{\text{base}} + \lambda \Delta_\ell$. The difficulty is that different base methods produce profiles with different magnitudes of departure from uniformity, so a $\lambda$ that meaningfully corrects one base may either over- or under-correct another. This makes cross-method comparison of the hyperparameter unreliable. We adopt instead a magnitude-normalized parameterization. Let $\bar{s}^{\text{base}} = \tfrac{1}{L}\sum_\ell s_\ell^{\text{base}}$ denote the base profile's mean and $\|s^{\text{base}} - \bar{s}^{\text{base}}\|_2$ its centered $\ell_2$ norm (the energy by which the base profile departs from a uniform allocation at the same total sparsity). Given a hyperparameter $\beta \geq 0$, we set
\begin{equation}
\lambda = \beta \cdot \frac{\|s^{\text{base}} - \bar{s}^{\text{base}}\|_2}{\|\Delta\|_2},
\label{eq:beta}
\end{equation}
where $\beta$ is the ratio of correction energy to the base profile's non-uniformity energy, and is therefore directly comparable across base methods.

\label{sec:clipping}

Given $\Delta_\ell$ and $\lambda$, we form $\tilde{s}_\ell = s_\ell^{\text{base}} + \lambda \Delta_\ell$, clip to $[0, 1]$, and iteratively redistribute any clipping residual across layers that still have headroom until the total matches $\sum_\ell s_\ell^{\text{base}}$ exactly, as detailed in Algorithm~\ref{alg:correction}. We grid-search $\beta \in \{0.05, 0.10, 0.20, 0.50\}$ on WikiText-2 perplexity.

\begin{algorithm}[t]  
\caption{Inter-Layer Coupling-Aware Correction.}
\label{alg:correction}
\begin{algorithmic}[1]  
\Require
    base profile $\{s_l^{\text{base}}\}$ summing to $S$;
    drift ratios $\{r_l\}$ from Section~\ref{sec:coefficient};
    strength $\beta$.
\Ensure
    corrected profile $\{s_l^{\text{new}}\}$ with $\sum_l s_l^{\text{new}} = S$.

\State $u_l \gets \operatorname{clip}\bigl(\log r_l,\; -\log 3,\; +\log 3\bigr)$
\State $\bar{u} \gets \frac{1}{L} \sum_l u_l$
\State $\Delta_l \gets -(u_l - \bar{u})$ \Comment{zero-mean correction signal}
\State $\lambda \gets \beta \cdot \frac{\|s^{\text{base}} - \bar{s}^{\text{base}}\|_2}{\|\Delta\|_2}$ \Comment{magnitude-normalized}
\State $\tilde{s}_l \gets s_l^{\text{base}} + \lambda \Delta_l$
\State $\hat{s}_l \gets \operatorname{clip}(\tilde{s}_l,\, 0,\, 1)$
\While{$|\sum_l \hat{s}_l - S| > \epsilon$}
    \State $\text{res} \gets S - \sum_l \hat{s}_l$
    \State $F \gets \{\, l : \hat{s}_l \text{ has headroom in the sign of res} \,\}$
    \State $\hat{s}_l \gets \operatorname{clip}\!\left(\hat{s}_l + \frac{\text{res}}{|F|} \cdot \mathbf{1}[l \in F],\; 0,\; 1\right)$
\EndWhile
\State \Return $\{\hat{s}_l\}$
\end{algorithmic}
\end{algorithm}

\begin{table*}[t]
\centering
\resizebox{\textwidth}{!}{%
\small
\setlength{\tabcolsep}{4pt}
\renewcommand{\arraystretch}{1.05}
\begin{tabular}{lccccccc}
\toprule
Method & LLaMA-2-7B & LLaMA-2-13B & LLaMA-3.2-3B & OPT-6.7B & Mistral-7B & Vicuna-7B & Qwen3-8B \\
\midrule
Dense              & 5.12            & 4.88              & 7.78            & 10.81               & 5.25             & 6.34             & 9.72            \\
\midrule
Wanda              & 72.90            & 46.61              & 134.55            & 160.00               & 60.34     & 59.47             & 82.53            \\
Wanda w.\ OWL       & 30.55            & 16.69              & 122.42            & 44.16               & 38.17             & 31.62             & 66.40            \\
Wanda w.\ OWL+abs   & \cellcolor{gray!20}\underline{23.19}{\scriptsize($\downarrow$24.09\%)}            & \cellcolor{gray!20}16.73{\scriptsize($\uparrow$0.24\%)}               & \cellcolor{gray!20}120.68{\scriptsize($\downarrow$1.42\%)}            & \cellcolor{gray!20}\textbf{43.03}{\scriptsize($\downarrow$2.56\%)}               & \cellcolor{gray!20}\underline{32.91}{\scriptsize($\downarrow$13.78\%)}             & \cellcolor{gray!20}31.26{\scriptsize($\downarrow$1.14\%)}             & \cellcolor{gray!20}64.38{\scriptsize($\downarrow$3.04\%)}            \\
Wanda w.\ AlphaPruning       & 31.76            & \textbf{15.20}              & 172.95            & 98.08       & 39.54             & 37.62             & 49.94            \\
Wanda w.\ AlphaPruning+abs   & \cellcolor{gray!20}29.98{\scriptsize($\downarrow$5.60\%)} & \cellcolor{gray!20}\underline{15.49}{\scriptsize($\uparrow$1.91\%)}  & \cellcolor{gray!20}166.19{\scriptsize($\downarrow$3.91\%)}            & \cellcolor{gray!20}94.70{\scriptsize($\downarrow$3.45\%)}                & \cellcolor{gray!20}33.89{\scriptsize($\downarrow$14.29\%)}             & \cellcolor{gray!20}\textbf{28.11}{\scriptsize($\downarrow$25.28\%)}             & \cellcolor{gray!20}\textbf{47.10}{\scriptsize($\downarrow$5.69\%)} \\
Wanda w.\ MRP       & 26.77   & 16.70     & \textbf{101.20} & 180.46                & 35.04            & 38.71    & 57.17            \\
Wanda w.\ ATP       & \textbf{22.00}            & 16.94              & \underline{103.81}   & \underline{43.69}    & \textbf{29.00} & \underline{28.13} & \underline{47.22}            \\
\midrule
SparseGPT          & 23.69            & 18.39              & 66.57            & \textbf{20.78}               & 23.12    & 30.48             & 107.59            \\
SparseGPT w.\ OWL  & 19.72            & 13.99              & 53.61            & 22.59               & 18.54             & 24.58    & 32.12            \\
SparseGPT w.\ OWL+abs & \cellcolor{gray!20}\textbf{18.27}{\scriptsize($\downarrow$7.35\%)}          & \cellcolor{gray!20}13.59{\scriptsize($\downarrow$2.86\%)}              & \cellcolor{gray!20}\textbf{52.71}{\scriptsize($\downarrow$1.68\%)}            & \cellcolor{gray!20}\underline{21.78}{\scriptsize($\downarrow$3.59\%)}               & \cellcolor{gray!20}\textbf{17.40}{\scriptsize($\downarrow$6.15\%)}             & \cellcolor{gray!20}23.78{\scriptsize($\downarrow$3.25\%)}             & \cellcolor{gray!20}\underline{29.68}{\scriptsize($\downarrow$7.60\%)}            \\
SparseGPT w.\ AlphaPruning   & 21.06 & \textbf{13.49}            & 74.10 & 76.33               & 20.58             & 26.85             & 33.08 \\
SparseGPT w.\ AlphaPruning+abs & \cellcolor{gray!20}19.93{\scriptsize($\downarrow$5.37\%)} & \cellcolor{gray!20}13.86{\scriptsize($\uparrow$2.74\%)}    & \cellcolor{gray!20}72.32{\scriptsize($\downarrow$2.40\%)}   & \cellcolor{gray!20}54.36{\scriptsize($\downarrow$28.78\%)}                & \cellcolor{gray!20}19.63{\scriptsize($\downarrow$4.62\%)} & \cellcolor{gray!20}\textbf{22.12}{\scriptsize($\downarrow$17.62\%)}    & \cellcolor{gray!20}29.80{\scriptsize($\downarrow$9.92\%)}   \\
SparseGPT w.\ MRP  & 20.13            & 13.99  & 53.45   & 35.47       & 18.96 & 25.83 & \textbf{26.95}            \\
SparseGPT w.\ ATP  & \underline{18.28}            & \underline{13.58}  & \underline{53.22}   & 24.27       & \underline{17.42} & \underline{23.23} & 29.77            \\
\bottomrule
\end{tabular}
}
\caption{WikiText-2 perplexity ($\downarrow$) of pruned models across architectures at 70\% sparsity. \textbf{Bold} marks the best result within each base-method block; \underline{underline} marks the second best. Lower is better.}
\label{tab:main-ppl}
\end{table*}

\section{Experiments}
\subsection{Experimental Settings.}
\textbf{Models.}
We evaluate our method across various LLMs, including LLaMA2 (7B/13B) \citep{touvron2023llama2}, LLaMA3.2-3B \cite{grattafiori2024llama}, OPT-6.7B \citep{zhang2022opt}, Vicuna-7B \citep{chiang2023vicuna}, Mistral-7B \citep{jiang2023mistral7b}, and Qwen3-8B \citep{yang2025qwen3}. 
Limited by available compute, we cap model sizes at 13B parameters, use FP16 arithmetic, and set the sequence length to 2048.
All experiments were conducted on a single NVIDIA L40S GPU (48GB).

\noindent \textbf{Evaluation.}
We evaluate language modeling performance via perplexity on the validation sets of WikiText-2 \citep{merity2016pointer}, and evaluate zero-shot performance on seven downstream tasks, including RTE \citep{wang2018glue}, WinoGrande \citep{sakaguchi2021winogrande}, BoolQ \citep{clark2019boolq}, OpenBookQA \citep{mihaylov2018can}, ARC-Easy/Challenge \citep{clark2018think}, and HellaSwag \citep{zellers2019hellaswag} from EleutherAI LM Harness \citep{eval-harness}.

\noindent \textbf{Pruning methods.}
We use Wanda \citep{sun2024simple} and SparseGPT \citep{frantar2023sparsegpt} as the underlying pruning procedures. We apply absorption-aware correction on top of two metric-based allocation methods: OWL \citep{yin2024outlier} and AlphaPruning \citep{lu2024alphapruning}. These methods score each layer from local channel-level signals, activation outliers and heavy-tailed weight spectra respectively, and are silent about inter-layer coupling, making them the natural targets for a correction that supplies this missing dimension. In addition, they themselves have good performance. For broader comparison, we additionally report results from a Uniform baseline and two representative non-uniform methods: MRP \citep{gao2025maximum}, which discovers allocations by iteratively pruning to balance layerwise redundancy; ATP \citep{huang2025determining}, which scores layers via an increasing arithmetic progression.

\noindent \textbf{Implementation Details.}
For calibration, we use 128 calibration samples from C4 \citep{raffel2020exploring} with sequence length 2048 as calibration dataset. All methods are evaluated with this identical dataset, the same pruning backbones, and comparable hyperparameters.

\subsection{Main Results.}

\textbf{Correction improves metric-based baselines.} 
Table \ref{tab:main-ppl} presents the perplexity of several LLM pruning methods and model families at 70\% sparsity on the WikiText-2 dataset, and additional results across different sparsity ratios are reported in Appendix \ref{app:all} . The results show that absorption-aware correction reduces perplexity of OWL and AlphaPruning on most model-sparsity combinations. Moreover, methods with absorption-aware correction achieve lower perplexity in most cases.

\begin{table*}[t]
\centering
\resizebox{\textwidth}{!}{%
\small
\setlength{\tabcolsep}{4pt}
\renewcommand{\arraystretch}{1.05}
\begin{tabular}{lccccccc}
\toprule
Method & LLaMA-2-7B & LLaMA-2-13B & LLaMA-3.2-3B & OPT-6.7B & Mistral-7B & Vicuna-7B & Qwen3-8B \\
\midrule
Dense              & 59.69            & 63.05              & 57.24            & 51.58               & 64.37             & 60.36             & 65.74            \\
\midrule
Wanda              & 33.43            & 37.44              & 34.13            & 36.72               & 36.85             & 36.96             & \textbf{43.03}   \\
Wanda w.\ OWL       & 40.50            & 47.44              & 36.60            & 39.34               & 39.25             & 44.08             & 38.82            \\
Wanda w.\ OWL+abs   & \cellcolor{gray!20}42.78{\scriptsize($\uparrow$5.63\%)}            & \cellcolor{gray!20}39.24{\scriptsize($\downarrow$17.28\%)}              & \cellcolor{gray!20}\underline{36.99}{\scriptsize($\uparrow$1.07\%)}            & \cellcolor{gray!20}\textbf{40.73}{\scriptsize($\uparrow$3.53\%)}               & \cellcolor{gray!20}39.57{\scriptsize($\uparrow$0.82\%)}             & \cellcolor{gray!20}44.91{\scriptsize($\uparrow$1.88\%)}             & \cellcolor{gray!20}39.24{\scriptsize($\uparrow$1.08\%)}            \\
Wanda w.\ AlphaPruning       & 42.06            & 47.30              & 33.06            & 39.39       & 38.96             & 43.33             & 40.27            \\
Wanda w.\ AlphaPruning+abs   & \cellcolor{gray!20}42.56{\scriptsize($\uparrow$1.19\%)} & \cellcolor{gray!20}47.55{\scriptsize($\uparrow$0.53\%)}  & \cellcolor{gray!20}33.61{\scriptsize($\uparrow$1.66\%)}            & \cellcolor{gray!20}40.27{\scriptsize($\uparrow$2.23\%)}               & \cellcolor{gray!20}39.79{\scriptsize($\uparrow$2.13\%)}             & \cellcolor{gray!20}\textbf{47.37}{\scriptsize($\uparrow$9.32\%)}             & \cellcolor{gray!20}41.00{\scriptsize($\uparrow$1.81\%)} \\
Wanda w.\ MRP       & \underline{43.45}   & \textbf{50.08}     & \textbf{37.07} & 39.29               & \underline{40.74}             & 45.53    & 38.54            \\
Wanda w.\ ATP       & \textbf{43.81}            & \underline{48.62}              & 36.54   & \underline{40.66}    & \textbf{41.08} & \underline{46.01} & \underline{42.10}            \\
\midrule
SparseGPT          & 41.99            & 45.28              & 38.36            & \underline{44.37}               & 44.02    & 43.19             & 39.48            \\
SparseGPT w.\ OWL  & 45.28            & 49.75              & 39.21            & 43.77               & 45.49             & \underline{46.93}    & 44.43            \\
SparseGPT w.\ OWL+abs & \cellcolor{gray!20}\underline{45.57}{\scriptsize($\uparrow$0.64\%)}          & \cellcolor{gray!20}49.67{\scriptsize($\downarrow$0.16\%)}              & \cellcolor{gray!20}39.40{\scriptsize($\uparrow$0.48\%)}            & \cellcolor{gray!20}\textbf{44.77}{\scriptsize($\uparrow$2.28\%)}               & \cellcolor{gray!20}45.23{\scriptsize($\downarrow$0.57\%)}
             & \cellcolor{gray!20}44.60{\scriptsize($\downarrow$4.96\%)}             & \cellcolor{gray!20}46.74{\scriptsize($\uparrow$5.20\%)}            \\
SparseGPT w.\ AlphaPruning   & 44.68 & 49.10            & 37.80 & 41.20               & 43.12             & 46.37             & 44.19 \\
SparseGPT w.\ AlphaPruning+abs & \cellcolor{gray!20}44.89{\scriptsize($\uparrow$0.47\%)} & \cellcolor{gray!20}48.97{\scriptsize($\downarrow$0.26\%)}    & \cellcolor{gray!20}38.03{\scriptsize($\uparrow$0.61\%)}   & \cellcolor{gray!20}42.55{\scriptsize($\uparrow$3.28\%)}               & \cellcolor{gray!20}42.82{\scriptsize($\downarrow$0.70\%)} & \cellcolor{gray!20}\textbf{47.22}{\scriptsize($\uparrow$1.83\%)}    & \cellcolor{gray!20}\underline{47.79}{\scriptsize($\uparrow$8.15\%)}   \\
SparseGPT w.\ MRP  & 45.41            & \textbf{51.87}  & \textbf{40.24}   & 42.59       & \underline{46.10} & 45.64 & 45.01            \\
SparseGPT w.\ ATP  & \textbf{46.03}            & \underline{50.25}  & \underline{39.76}   & 43.35       & \textbf{46.47} & 46.80 & \textbf{47.85}            \\
\bottomrule
\end{tabular}
}
\caption{Seven zero-shot tasks' average accuracy ($\uparrow$) of pruned models across architectures at 70\% sparsity. \textbf{Bold} marks the best result within each base-method block; \underline{underline} marks the second best. Higher is better.}
\label{tab:main-acc}
\end{table*}

\noindent \textbf{Zero-Shot Evaluation.}
Table \ref{tab:main-acc} presents accuracy results obtained from seven zero-shot tasks at 70\% sparsity. Aligned with perplexity result, methods with absorption-aware correction regularly deliver the best or second-best performance among non‑uniform pruning alternatives when paired with identical base pruning algorithms across diverse model architectures.

\noindent \textbf{Computational Overhead of Absorption Coefficient Estimation.}
Because the gains reported above come from a signal that requires forward passes beyond what baseline allocation methods need, we report the additional computational cost incurred by absorption-aware correction.

\begin{table}[h]
\centering
\small
\setlength{\tabcolsep}{4pt}
\caption{Computational overhead of absorption coefficient estimation on LLaMA-2 models. $T_{\mathrm{prune}}$ and $T_{\mathrm{abs}}$ denote the wall-clock time, in minutes, of base sparseGPT pruning and absorption coefficient estimation, respectively.}
\label{tab:overhead}
\begin{tabular}{lccc}
\toprule
Model & $T_{\mathrm{prune}}$ & $T_{\mathrm{abs}}$ & $T_{\mathrm{abs}}/T_{\mathrm{prune}}$ \\
\midrule
LLaMA-2-7B  & 6.85 & 1.90 & 27.7\% \\
LLaMA-2-13B & 11.47 & 4.08 & 35.6\% \\
\bottomrule
\end{tabular}
\end{table}

Table \ref{tab:overhead} reports wall-clock time on a single NVIDIA L40S GPU (48GB). 
Absorption estimation takes 1.90 minutes per 7B model, which represents 27.7\% of the sparsegpt pruning time. We conclude that absorption-aware correction is practical to add to existing pruning pipelines without substantially increasing the computational cost.

\subsection{Ablation Studies}

To verify that the improvement does not merely come from introducing an additional searchable correction direction, we replace the absorption-derived correction vector with a random Gaussian vector. For each seed, we sample $z_\ell \sim \mathcal{N}(0,1)$ and construct a centered random correction direction which is then normalized following the same procedure used for the real absorption-based correction vector. We search its correction strength over the same range. The resulting sparsity profile is processed with the same clipping and budget-preserving redistribution procedure, ensuring an unchanged global sparsity budget. We repeat this experiment over multiple random seeds and report the mean in Table~\ref{tab:Gaussian_pruning}. As shown in the table, the random correction vector does not provide improvements comparable to the absorption-derived correction, indicating that the gains are attributable to the layer-specific absorption information encoded in the proposed correction direction rather than to the extra searchable degree of freedom introduced by the correction framework.

\begin{table}[t]
\centering
\caption{WikiText-2 perplexity of Gaussian random correction ablation on LLaMA-2-7B with Wanda under different pruning ratios.}
\label{tab:Gaussian_pruning}
\begin{tabular}{lcc}
\toprule
Method & 60\% & 70\% \\
\midrule
OWL             & 9.18 & 30.55 \\
OWL+abs              & 9.02 & 23.19 \\
OWL+Gaussian          & 9.17 & 36.72 \\
AlphaPruning     & 9.81 & 31.76 \\
AlphaPruning+abs     & 9.18 & 29.98 \\
AlphaPruning+Gaussian & 9.90 & 32.96 \\
\bottomrule
\end{tabular}
\end{table}

\section{Conclusion}
We studied layer-wise sparsity allocation through a perturbation-response perspective, asking not which layers are intrinsically important but how the rest of the network responds to the perturbations that pruning introduces. Through controlled injection experiments, we found that transformer layers exhibit highly heterogeneous responses to pruning-scale perturbations and that this absorption is a nonlinear, large-perturbation phenomenon that emerges at the scale of actual pruning. Building on these findings, we defined an absorption coefficient that measures this property directly, and showed that applying absorption-aware correction to OWL and AlphaPruning improves their performance in most settings and yields clear average gains. We hope these findings encourage a shift in layer-wise pruning research toward direct characterization of how pruning perturbations propagate, rather than further refinement of proxy signals for layer importance.

\bibliography{latex/custom}

\end{document}